\def\year{2019}\relax
\documentclass[letterpaper]{article} 
\usepackage{aaai19}  
\usepackage{times}  
\usepackage{helvet}  
\usepackage{courier}  
\usepackage{url}  
\usepackage{graphicx}  
\usepackage{amsmath,amssymb} 
\usepackage{color}
\usepackage{enumitem}
\usepackage{tabularx}
\usepackage{multirow}
\usepackage{sidecap}
\usepackage[lined,boxed,commentsnumbered]{algorithm2e}
\nocopyright  
 
\begin{document}
%
\title{Towards Interpretable R-CNN by Unfolding Latent Structures}
\author{Tianfu Wu$^{1,2}$, Wei Sun$^1$, Xilai Li$^1$, Xi Song, and Bo Li$^3$\\
Department of ECE$^1$ and Visual Narrative Initiative$^2$, 
North Carolina State University\\
YunOS BU, Alibaba Group$^3$\\
{\tt\small \{tianfu\_wu, xlai47, wsun12\}@ncsu.edu, \{xsong.lhi, boli.lhi\}@gmail.com}}
\maketitle
\begin{abstract}
This paper first proposes a method of formulating model interpretability in visual understanding tasks based on the idea of unfolding latent structures. It then presents a case study in object detection using popular two-stage region-based convolutional network (i.e., R-CNN) detection systems~\cite{FastRCNN,FasterRCNN,RFCN,MaskRCNN}.  We focus on weakly-supervised extractive rationale generation, that is learning to unfold latent discriminative part configurations of object instances automatically and simultaneously in detection without using any supervision for part configurations. We utilize a top-down hierarchical and compositional grammar model embedded in a directed acyclic AND-OR Graph (AOG) to explore and unfold the space of latent part configurations of regions of interest (RoIs). We propose an AOGParsing operator to substitute the RoIPooling operator widely used in R-CNN. In detection, a bounding box is interpreted by the best parse tree derived from the AOG on-the-fly, which is treated as the qualitatively extractive rationale generated for interpreting detection. We propose a folding-unfolding method to train the AOG and convolutional networks end-to-end. In experiments, we build on R-FCN~\cite{RFCN} and test our method on the PASCAL VOC 2007 and 2012 datasets. We show that the method can unfold promising latent structures without hurting the performance. 
\end{abstract}

\section{Introduction}
Recently, deep neural networks~\cite{LeCunCNN,AlexNet} have improved prediction accuracy significantly in many vision tasks, and even outperform humans in image classification tasks~\cite{ResidualNet,InceptionNet}. In the literature of object detection, there has been a critical shift from more explicit representation and models such as the mixture of deformable part-based models (DPMs)~\cite{DPM} and its many variants, and hierarchical and compositional AND-OR graphs (AOGs) models~\cite{DisAOT-CVPR,Yuille_AndOr,carAOG_PAMI,TLP-PAMI}, to less transparent but much more accurate ConvNet based approaches~\cite{FasterRCNN,RFCN,YOLO,SSD,MaskRCNN,DCN}.
Meanwhile, it has been shown that deep neural networks can be easily fooled by so-called adversarial attacks which utilize visually imperceptible, carefully-crafted perturbations to cause networks to misclassify inputs in arbitrarily chosen ways~\cite{FoolDeepNet,AdversarialExampl}, even with one-pixel attack~\cite{OnePixelAttack}. And, it has also been shown that deep learning can easily fit random labels~\cite{GeneralizationOfDL}.  It is difficult to analyze why state-of-the-art deep neural networks work or fail due to the lack of theoretical underpinnings at present~\cite{BoundsOfDL}. From cognitive science perspective, state-of-the-art deep neural networks might not learn and think like people who  know and can explain ``why"~\cite{BuildMachineLikePeople}.   Nevertheless, there are more and more applications in which prediction results of computer vision and machine learning modules based on deep neural networks have been used in making decisions with potentially critical consequences (e.g., security video surveillance  and autonomous driving). 

 \begin{figure*} [t]
   \centering
   {\includegraphics[width = 1\textwidth]{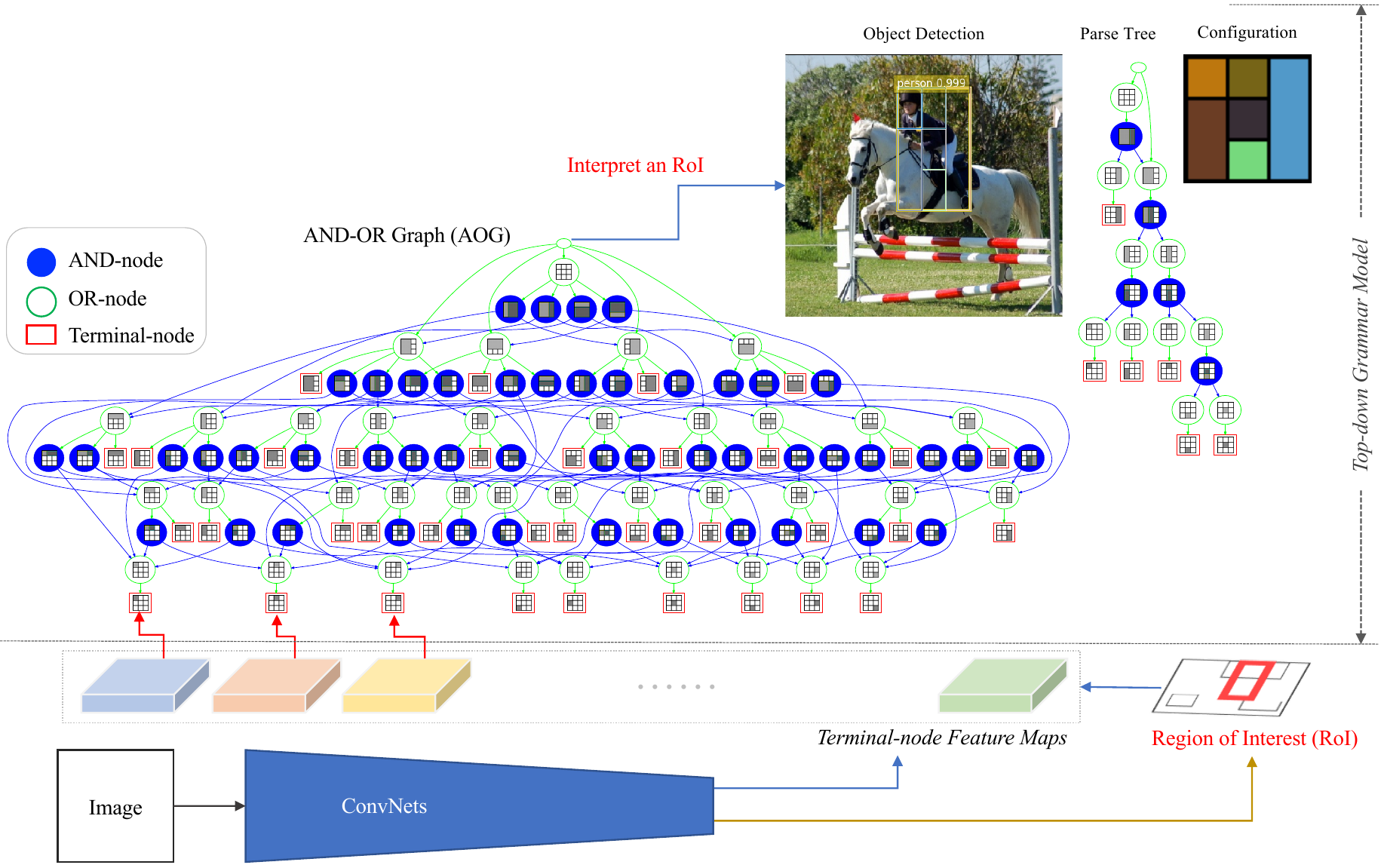}}
   \caption{Illustration of the proposed end-to-end integration of a generic top-down grammar model represented by a directed acyclic AND-OR Graph (AOG) and bottom-up ConvNets. For clarity, we show an AOG constructed for a $3\times 3$ grid using the method proposed in~\cite{DisAOT-CVPR}. The AOG unfolds the space of all possible latent part configurations. We build on the R-FCN method~\cite{RFCN}. Based on the AOG, we use Terminal-node sensitive maps and propose an AOGParsing operator to substitute the position-sensitive RoIPooling operator in the R-FCN, which will infer the best parse tree for a RoI on-the-fly, as well as the best part configuration.  See text for details. (Best viewed in color and magnification) 
   }
   \label{fig:model}
   \end{figure*} 
 
It has become a common recognition that prediction without interpretable justification will have limited applicability eventually. For example, consider the intuitive fact that people could get frustrated if someone close to her/him did something critical without convinced explanation, let alone machine systems. So, it is a crucial issue of addressing machine's inability to explain its predicted decisions and actions (e.g., eXplainable AI or XAI proposed in the DARPA grant solicitation~\cite{XAI}), that is to improve accuracy and transparency jointly: Not only is an interpretable model capable of computing correct predictions of a random example with very high probability, but also rationalizing its predictions, preferably in a way explainable to end users. Generally speaking, learning interpretable models is to let machines make sense to humans, which usually consists of many challenging aspects. So there has not been a universally accepted definition of the notion of model interpretability. Especially, it remains a long-standing open problem to measure interpretability in a quantitative and principled way. 

To address the explainability challenge, many work have proposed to visualize the internal filter kernels or to generate attentive activation maps, which reveal a lot of insights of what DNNs have learned in a post-hoc way. Complementary to those methods, this paper focuses on how to unfold the latent structures for addressing model interpretability in learning and inference. We first propose a method of formulating model interpretability. We then present a case study in object detection. We aim to investigate the feasibility of integrating a top-down model representing the space of latent structures with ConvNets end-to-end, and to qualitatively rationalize the popular  two-stage region-based ConvNets detection system, i.e.  R-CNN~\cite{FastRCNN,FasterRCNN,RFCN} without hurting the detection performance. 

 Figure~\ref{fig:model} illustrated the proposed method for object detection. It integrates a generic top-down hierarchical and compositional grammar model and bottom-up ConvNets end-to-end. We adopt  R-CNN~\cite{FastRCNN,FasterRCNN,RFCN} in detection. We focus on weakly-supervised extractive rationale generation in the RoI prediction component in R-CNN, that is learning to unfold latent discriminative part configurations of RoIs automatically and simultaneously in detection without using any supervision for part configurations. 
 To that end, we utilize a generic top-down hierarchical and compositional grammar model embedded in a directed  acyclic AND-OR Graph (AOG)~\cite{DisAOT-CVPR,TLP-PAMI} to explore and unfold the space of latent part configurations of RoIs (see an example in the top of Figure~\ref{fig:model}). There are three types of nodes in an AOG: an \textit{AND-node} represents binary decomposition of a large part into two smaller ones, an \textit{OR-node} represents alternative ways of decomposition, and a \textit{Terminal-node} represents a part instance. 
 The AOG is consistent with  the general image grammar framework~\cite{Geman_CompositionSystems,Zhu_Grammar,Pff_Grammar,Yuille_AndOr}.  
 We propose an \textbf{AOGParsing operator} to substitute the RoIPooling operator in the R-CNN based detection systems. In detection, each bounding box is  interpreted by the best parse tree derived from the AOG on-the-fly, which is the extractive rationale generated for detection. 
 
 
 In experiments, we build on the R-FCN~\cite{RFCN} with the residual net~\cite{ResidualNet} pretrained on the ImageNet~\cite{ImageNet} as backbone. We test our method on the PASCAL VOC 2007 and 2012 datasets with performance comparable to state-of-the-art methods. We also perform the ablation study on different aspects of the proposal method. 

 \section{Related Work}

In the literature, many work focused on interpreting post-hoc interpretability of deep neural networks by associating explanatory semantic information with nodes in a deep neural network. 
There are a variety of methods including identifying high-scoring image patches~\cite{GirshickRCNN,visConvnet} or over-segmented atomic regions~\cite{RibeiroSG16} directly, visualizing the layers of convolutional networks using deconvolutional networks to understand what contents are emphasized in the high-scoring input image patches~\cite{deconvolutionalnetwork}, identifying items in a visual scene and recount multimedia events~\cite{visCNN,devnet}, generating synthesized images by maximizing the response of a given node in the network~\cite{visHighOrderFeatures,visHighOrderFeatures1,visConvent1} or by developing a top-down generative convolutional networks~\cite{cnnFrame,GAN-cnn}. Hendricks et al~\cite{GenerateExplanations} extended the approaches used to generate image captions~\cite{ImageCaptioning,ImageCaptioning1} to train a second deep network to generate explanations without explicitly identifying the semantic features of the original network. Most of these methods are not model-agnostic except for~\cite{RibeiroSG16}. 

More recently, Spatial attention-like mechanism has been widely studied in deep neural network based systems, including the seminal spatial transform network~\cite{STN} which warps the feature map via a global parametric transformation such as affine  transformation, the exploration of global average pooling and class specific activation maps for weakly-supervised discriminative localizationj(i.e., CAM)~\cite{DisLocalization}, the deformable convolution network~\cite{DCN} and active convolution~\cite{ActiveConv}, and more explicit attention based work in image caption and visual question answering (VQA) such as the show-attend-tell work~\cite{ShowAttendTell} and the hierarchical co-attention in VQA~\cite{coAttentionVQA}.
The Grad-CAM work~\cite{GradCAM}, extended from the CAM work~\cite{CAM}, can produce a coarse localization map highlighting the important regions in the image used by deep neural networks for predicting the concept. In similar spirit, the excitation back-propagation method~\cite{ExcitationBackprop} can generate task-specific attention map. The latest network dissection work~\cite{netdissect2017} reported empirically that interpretable units are found in representations of the major deep learning architectures~\cite{AlexNet,VGG,ResidualNet} for vision, and interpretable units also emerge under different training conditions. On the other hand, they also found that interpretability is neither an inevitable result of discriminative power, nor is it a prerequisite to discriminative power.  Most of these methods are not model-agnostic except for~\cite{RibeiroSG16,koh2017understanding}. In~\cite{koh2017understanding}, a classic technique in statistics, influence function, is used to understand the black-box prediction in terms of training sample, rather than extractive rationale justification.

\textbf{Our Contributions.} This paper makes three main contributions to the emerging field of learning interpretable models as follows: (i) It presents a method of integrating a generic top-down grammar model, embedded in an AOG, and bottom-up ConvNets end-to-end to learn qualitatively interpretable models in object detection. 
(ii) It presents an AOGParsing operator which can be used to substitute the RoIPooling operator widely used in R-CNN based detection systems. 
(iii) It shows detection performance comparable to state-of-the-art R-CNN systems, thus shedding light on addressing accuracy and transparency jointly in learning deep models for object detection.

\section{Interpreting Model Interpretability}
In this section, we present a generic formulation of model interpretability in visual understanding tasks which accounts for unfolding well-defined latent structures in a weakly-supervised way. 

Intuitively, we would expect that an interpretable model could learn and capture latent semantic structures automatically which are not annotated in training data. For example, if we consider the basic image classification task with only image labels available in training as commonly used, to compare which classification models are more interpretable or explainable, one principled way is to show the capability of extracting the latent localization of object of interest w.r.t. the ground-truth label.  Similarly, a person detector is more interpretable if it is learned using person bounding box annotations only, but capable of interpreting a person detection with the latent semantic structure explained, ideally the kinetic pose. So, \textit{our intuitive idea is that model interpretability can be posed as the capability of exploring the latent space of a higher level task} (e.g., localization vs classification and pose recovery vs detection) in a principled way, and of capturing the sufficient statistics in the latent space.  
The more a model can explore and capture the latent tasks at higher level, the better the model interpretability is.

To that end, we first consider an underlying task hierarchy, e.g., from image classification, to object localization and detection, to object part recovery (object parsing), and all the way to full image parsing (i.e., all image pixels are explained-away in a mathematically sound way). Then, for a task at hand (e.g., object detection), we seek a principled way of defining and exploring the latent space of the task of object part-based parsing, and then compute extractive rationale for the task at hand. 

Our formulation is a straightforward top-down method. We first build a grammar structure which quantizes and unfolds the space of latent structures by  utilizing the methods presented in~\cite{DisAOT-CVPR,TLP-PAMI,Tangram_TIP}. Then, we integrate the grammar structure into the model in learning and inference. The parse graph of the grammar structure is treated as the qualitatively interpretable result. The grammar structure can be potentially exploited to build quantitatively interpretable models from scratch by defining loss functions on the latent structures captured by the grammar. To investigate the feasibility, we present a case study on object detection in this paper. 

\section{A Case Study: Toward Interpretable R-CNN}\label{sec:formulation}
In this section, we first briefly present backgrounds on R-CNN  and the construction of the top-down AOG~\cite{DisAOT-CVPR,TLP-PAMI} to be self-contained. Then, we present the end-to-end integration of AOG and ConvNets. 

\textbf{The R-CNN Framework.}
 The R-CNN framework consists of three components: (i) A ConvNet backbone such as the Residual Net~\cite{ResidualNet} for feature extraction, parameterized by $\Theta_0$ and shared between the region-proposal network (RPN) and the RoI prediction network. (ii) The RPN network for objectness detection (i.e., category-agnostic detection through binary classification between foreground objects and background) and bounding box regression, parameterized by $\Theta_1$. Denote by $B$ a RoI (i.e., a foreground bounding box proposal) computed by the RPN. (iii) The RoI prediction network for classifying a RoI $B$ and refining it, parameterized by $\Theta_2$, which utilizes the RoIPooling operator and usually use one or two fully connected layer(s) as the head classifier and regressor. We build on top of the R-FCN method~\cite{RFCN} in our experiments. In R-FCN, position-sensitive score maps are used in RoIPooling, that is to treat cells in a RoI as object parts each of which has its own score map. The final classification is based on the majority voting after RoIPooling.  The parameters $\Theta=(\Theta_0, \Theta_1, \Theta_2)$ are trained end-to-end. 

\textbf{The AOG.}
In the R-CNN framework, a RoI is interpreted as a predefined flat configuration. To learn interpretable models, we need to explore the space of latent part configurations defined in a RoI. To that end, a RoI is first divided into a grid of cells as done in the RoIPooling operator  (e.g., $3\times 3$ or $7\times 7$). Denote by $S_{x,y,w,h}$ and $t_{x,y,w,h}$ a non-terminal symbol and a terminal symbol respectively, both representing the sub-grid with left-top $(x,y)$ and width and height $(w,h)$ in the RoI. We only utilize binary decomposition, either $Hor$izontal cut  or $Ver$tical cut, when interpreting a non-terminal symbol. We have four rules, 
\begin{align}
S_{x,y,w,h}  &\xrightarrow{Termination}  t_{x,y,w,h} \\
S_{x,y,w,h}(l;\leftrightarrow)  &\xrightarrow{Ver.Cut}  S_{x, y, l, h} \cdot S_{x+l, y, w-l, h}\\
S_{x,y,w,h}(l;\updownarrow)  &\xrightarrow{Hor.Cut}  S_{x, y, w, l} \cdot S_{x, y+l, w, h-l}\\
S_{x,y,w,h}  &\rightarrow  t_{x,y,w,h} | \\
\nonumber & S_{x,y,w,h}(l_{min};\leftrightarrow) | \cdots | S_{x,y,w,h}(w-l_{min};\leftrightarrow) |\\
\nonumber & S_{x,y,w,h}(l_{min};\updownarrow) | \cdots | S_{x,y,w,h}(h-l_{min}; \updownarrow)，
\end{align}
where $l_{min}$ represents the minimum side length of a valid sub-grid allowed in the decomposition (e.g., $l_{min}=1$). 
When instantiated, the first rule will be represented by \textit{Terminal-nodes}, both the second and the third  by \textit{AND-nodes}, and the fourth by \textit{OR-nodes}.

The top-down AOG is constructed by applying the four rules in a recursive way~\cite{DisAOT-CVPR,TLP-PAMI}. Denote an AOG by $
  \mathcal{G} = (V, E)$
  where $V=V_{And}\cup V_{Or}\cup V_{T}$ and $V_{And}, V_{Or}$ and $V_T$ represent a set of AND-nodes, OR-nodes and Terminal-nodes respectively, and $E$ a set of edges. We start with $V=\emptyset$ and $E=\emptyset$, and a first-in-first-out queue $Q=\emptyset$. 
It unfolds all possible latent configurations. We further introduce a super OR-node whose child nodes are those OR-nodes that occupy the entire grid more than certain threshold (e.g., $0.5$). The super OR-node is used in the unfolding step of learning the AOG model to help find better interpretation for noisy RoIs from the RPN network.  
Figure~\ref{fig:model} shows the AOG constructed for a $3\times 3$ grid. In~\cite{DisAOT-CVPR}, The two child nodes of an AND-node are allowed to overlap up to certain ratio, which we do not use in our experiments for simplicity. 

\textit{A parse tree} is an instantiation of the AOG, which follows the breadth-first-search (BFS) order of nodes in the AOG, selects the best child node for each encountered OR-nodes, keeps both child nodes for each encountered AND-node, and terminates at each encountered Terminal-node. \textit{A configuration} is generated by collapsing all the Terminal-nodes of a parse tree onto the image domain. 


\subsection{The Integration of AOG and ConvNets.}
We now present a simple end-to-end integration of the top-down AOG and ConvNets, as illustrated in  Figure~\ref{fig:model}. 

Consider an AOG $\mathcal{G}_{h,w,l_{min}}$ with the grid size being $h\times w$ and the minimum side length $l_{min}$ allowed for nodes (e.g., $\mathcal{G}_{3,3,1}$ in Figure~\ref{fig:model}). Denote by $F(v)\in \mathbb{F}$ the Terminal-node sensitive score map for a Terminal-node $v\in V_{T}$. All $F(v)$'s have the same dimensions, $C\times H\times W$, where the height $H$ and the width $W$ are the same as those of the last layer in the ConvNet backbone, and the channel $C$ the number of classes in detection (e.g., $C=21$ in the PASCAL VOC benchmarks including $20$ foreground categories and $1$ background). $F(v)$'s are usually computed through $1\times 1$ convolution on top of the last layer in the ConvNet backbone.  Denote by $f_B(v)$ the $C$-d score vector of a Terminal node $v$ placed in a RoI $B$, which is computed by average pooling in the corresponding sub-grid occupied by $v$ (in the same way that the position-sensitive RoIPooling of R-FCN computes the score vector of a RoI grid cell). Following the depth-first search (DFS) order, the score vectors of Terminal-nodes are then passing through the AOG in the forward step w.r.t. the folding-unfolding stage in learning. Following the breadth-first-search (BFS) order, the best parse tree per category for a RoI is inferred in the backward step in the unfolding stage, as well as the part configuration. We elaborate the forward and backward computation in the next section. 

\textit{Remark:} The number of channels $C$ of Terminal-node score maps can take values other than the number of classes, and we add a fully connected layer on top of the AOG to predict the class scores. We keep it simple in this paper. 

\begin{algorithm} [t]
 \SetAlgoLined
 \underline{\textbf{Forward}$(\mathcal{G}, \mathbb{F}, B, isFolding)$}\;
 \While{$Q_{DFS}(\mathcal{G})$ is not empty}{
  Pop a node $v$ from the $Q_{DFS}(\mathcal{G})$\; 
  \uIf{$v$ is an OR-node}{
	  \uIf{$isFolding$}{
	  $f_B(v) = \sum_{u\in ch(v)} f_B(u) / |ch(v)|$
	  $\omega_0(v) = \sum_{u\in ch(v)} \omega_0(u) / |ch(v)|$
	  }
	  \Else {
  	$f_B(v) = \max_{u\in ch(v)}f_B(u)$ \small{(element-wise)}
  	$i_B(v) = \arg\max_{u\in ch(v)} f_B(u)$ 
  	}
  }
  \uElseIf{$v$ is an AND-node} {
  	$f_B(v) = \sum_{u\in ch(v)} f_B(u)$
  	$\omega_0(v) = \sum_{u\in ch(v)} \omega_0(u)$
  }
  \ElseIf{$v$ is a Terminal-node} {
  	Compute $f_B(v)$ in $F(v)\in \mathbb{F}$, and $\omega_0(v)=1$
  }  
 }
 
 Normalize the score vector of the root OR-node $O$\;
 \uIf{$isFolding$}{
 	  $f_B(O) = f_B(O) / \omega_0$
 	  }
 	  \Else {
 	  Compute $\omega_1$ using Algorithm~\ref{alg:w1}\;
 	  $f_B(O) = f_B(O) / \omega_1$  \small{(element-wise)}
 	  }
  \caption{Forward computation with AOG.}\label{alg:forwad} 
\end{algorithm}

\begin{algorithm} [!ht]
 \SetAlgoLined
 \underline{\textbf{ComputeOmega1}$(\mathcal{G}, i_B)$}\;
 \For{$c=0, \cdots, C-1$} {
 Initialize $q_{BFS}=\{O\}$ ($O$ is the root OR-node) \;
 Set $\omega_1(c)=0$\;
 \While{$q_{BFS}$ is not empty}{
  Pop a node $v$ from the $q_{BFS}$\; 
  \uIf{$v$ is an OR-node}{
	  Push the best child node in $q_{BFS} = q_{BFS}\cup \{i_B(v,c)\}$
  }
  \uElseIf{$v$ is an AND-node} {
  	Push the two child nodes in $q_{BFS} = q_{BFS}\cup ch(v)$
  }
  \ElseIf{$v$ is a Terminal-node} {
  	$\omega_1(c) += 1$
  }  
 }
 }
  \caption{Computing the forward normalization weight $\omega_1$ in the unfolding stage.}\label{alg:w1}
\end{algorithm}  

\begin{algorithm} [t]
 \SetAlgoLined
 \underline{\textbf{Backward}$(\mathcal{G}, \mathbf{g}, B, isFolding)$}\;
 \uIf{$isFolding$}{
   	  $\mathbf{g} = \mathbf{g} / \omega_0$ 
   	  where $\omega_0$ is computed in Algorithm~\ref{alg:forwad}\;
   	  }
   	  \Else {
   	  $\mathbf{g} = \mathbf{g} / \omega_1$ where $\omega_1$ is computed in Algorithm~\ref{alg:w1}\;
   	  }
  For the root OR-node $O$, we have  	  
  $g_B(O) = \mathbf{g}$

 \While{$Q_{BFS}(\mathcal{G})$ is not empty}{
  Pop a node $v$ from the $Q_{BFS}(\mathcal{G})$\; 
  \uIf{$v$ is an OR-node}{
	  \uIf{$isFolding$}{
	    	  	    	  
	    	  $g_B(u) += g_B(v), \quad \forall u\in ch(v)$
	    	  }
	    	  \Else {	    	  
	    	  
	    	  $g_B(u,c) += g_B(v,c), \quad \text{ if } i_B(v, c)=u, \forall u\in ch(v), c\in \{0,\cdots, C-1\}$
	    	  where $i_B()$ is computed in Algorithm~\ref{alg:forwad}\;
	    	  }
  }
  \uElseIf{$v$ is an AND-node} {
  	$g_B(u) += g_B(v), \quad \forall u\in ch(v)$
  }
  \ElseIf{$v$ is a Terminal-node} {
  	Back-propagate $g_B(v)$ to Terminal-sensitive feature maps $F(v)$.
  }  
 }
  
  \caption{Backward computation with AOG.}\label{alg:backward} 
\end{algorithm}  

\subsection{The Folding-Unfolding Learning}
Since the Terminal-node sensitive maps are computed with randomly initialized convolution kernels, it is not reasonable to select the best child for each OR-node at the beginning in the forward step, and all nodes not retrieved by the parse trees will not get gradient update in the backward step. So, we resort to a folding-unfolding learning strategy. In the folding stage, OR-nodes are implemented by MEAN operators and AND-nodes by SUM operators, thus the AOG is actually an AND Graph and all nodes will be updated in the backward computation. The folding stage is usually trained for one or two epochs. In the unfolding stage, OR-nodes are implemented by element-wise MAX operators and AND-nodes still by SUM operators, leading to the \textbf{AOGParsing operator}. For notational simplicity, we write $\mathcal{G}$ for an AOG $\mathcal{G}_{w,h,l_{min}}$. In both forward and backward computation, all RoIs are processed at once in implementation and we present the algorithms using one RoI $B$ for clarity. 

\textbf{Forward Computation.} Denote by $Q_{DFS}(\mathcal{G})$ the DFS queue of nodes in an AOG $\mathcal{G}$.  
Forward computation (see Algorithm~\ref{alg:forwad}) is to compute score vectors for all nodes following $Q_{DFS}(\mathcal{G})$ in both folding and unfolding stages. It also computes the assignment of the best child node  of OR-nodes in unfolding stage, denoted by $i_B(v), v\in V_{Or}$. In the forward step, the score vector of the root OR-node needs to be normalized for fair comparison, especially in the unfolding stage where different parse trees have different number of Terminal-nodes. Denote by $\omega_0(v)$ the normalization weight in the folding stage which is a scalar shared by all categories. Denote by and $\omega_1$ the normalization weight vector in the unfolding stage which is a $C$-d vector since different categories might infer different best parse trees in interpreting a RoI $B$.

\textbf{Backward Computation.} Similarly, by changing the DFS queue to the BFS queue, we can define backward computation using the AOG based on Algorithm~\ref{alg:forwad} for the folding stage, and on  Algorithm~\ref{alg:w1} for the unfolding stage which is summarized in Algorithm~\ref{alg:backward}. 


 \begin{table*} [t]
      \caption{Performance comparisons using Average Precision (AP) at the intersection over union (IoU) threshold $0.5$ (AP@$0.5$) and $0.7$ (AP@$0.7$) respectively in the PASCAL VOC2007 dataset (using the protocol, competition "comp4" trained using both 2007 and 2012 trainval datasets). In the table, ``\textbf{fAOG772-d}" represents the model trained using the deformable AOG $\mathcal{G}_{7,7,2}$ and the folding stage only, ``\textbf{AOG772-d-7 or AOG772-d-1}" the model trained using the folding-unfolding method with the unfolding stage initialized from the model at epoch $7$ or $1$ in the folding stage respectively. Without ``-d", it means the AOGs are not deformable. ``RFCN-d-re" represents the reproduced results of R-FCN with deformable convolution using our modified code which are consistent with the results reported in~\cite{DCN}.}\label{table:voc2007} 
      \centering 
      \resizebox{\textwidth}{!}{
      {\renewcommand{\arraystretch}{1.8}
      \tiny 
     \begin{tabular}{|l*{19}{|c}|}
     \hline 
     &
     \rotatebox[origin=c]{270}{fAOG772-d} & \rotatebox[origin=c]{270}{AOG772-d-7} & \rotatebox[origin=c]{270}{AOG772-d-1} & \rotatebox[origin=c]{270}{fAOG772}  &  \rotatebox[origin=c]{270}{AOG772-7} & \rotatebox[origin=c]{270}{AOG772-1}  & \rotatebox[origin=c]{270}{fAOG551-d} & \rotatebox[origin=c]{270}{AOG551-d-7} & \rotatebox[origin=c]{270}{AOG551-d-1} & \rotatebox[origin=c]{270}{fAOG551}  &  \rotatebox[origin=c]{270}{AOG551-7} &  \rotatebox[origin=c]{270}{AOG551-1} & \rotatebox[origin=c]{270}{fAOG331-d} & \rotatebox[origin=c]{270}{AOG331-d-7} & \rotatebox[origin=c]{270}{AOG331-d-1} & \rotatebox[origin=c]{270}{fAOG331}  &  \rotatebox[origin=c]{270}{AOG331-7} &  \rotatebox[origin=c]{270}{AOG331-1} & \rotatebox[origin=c]{270}{RFCN-d-re}  \\ \hline
     AP@$0.5$ & 81.7 & 80.7 & 82.0 & 80.2 & 81.1 & 81.1 & 81.5 & 80.7 & \bf{82.1} & 81.1 & 80.5 & 81.7 & 81.5 & 80.8 & 81.3 & 80.4 & 80.2 & 80.3 & 82.0 \\ \hline     
     AP@$0.7$ & 67.8 & \bf{68.1} & 68.6 & 66.1 & 67.7 & 66.7 & 66.9 & 68.4 & 67.9 & 67.4 & 67.9 & 66.8 & 67.4 & 67.8 & 67.1 & 66.7 & 67.7 & 67.0 & 67.9 \\ \hline
 	\end{tabular} 
 	} }
 \end{table*}  
 \begin{table*} [!t]
      \caption{Performance comparisons using AP@$0.5$ in the PASCAL VOC2012 dataset (``comp4"). ``AOG772-d-1" can be viewed at  \url{http://host.robots.ox.ac.uk:8080/anonymous/EXCJXR.html}, and ``RFCN-d-re" at \url{http://host.robots.ox.ac.uk:8080/anonymous/BWL8DV.html}. }\label{table:voc2012} 
      
      \resizebox{\textwidth}{!}{
           {\renewcommand{\arraystretch}{1.8}
           \tiny    
     \begin{tabular}{|l*{21}{|c}|}
     \hline 
     & aero & bike & boat & bttle & bus & car & mbik & train & bird & cat & cow & dog & hrse & sheep & pers & plant & chair & tble & sofa & tv & avg. \\ \hline
 	AOG772-d-1 & 87.7 & 84.1 & 79.5 & 66.7 & 63.3 &82.2 &81.3 &93.6 &61.2 &82.4 &62.2 &92.2 & 87.4 & 85.9 & 84.9 & 60.2 & 83.4 & 69.9 & 87.0 & 73.4 & \bf{78.4} \\ \hline
 	RFCN-d-re & 87.0 & 84.3 & 78.8 & 67.8 & 62.2 & 80.9 & 81.7 & 93.8 & 60.3 & 82.5 & 63.4 & 92.2 & 87.0 & 86.6 & 85.5 & 60.3 & 82.8 & 68.8 & 86.4 & 73.5 & 78.3 \\ \hline  
     \end{tabular} 
     } }
 \end{table*}
 
 \section{Experiments}\label{sec:exp}
 In this section, we present experimental results on the PASCAL VOC 2007 and 2012 benchmarks~\cite{VOC}. We also give the ablation study on different aspects of the proposed method. We build on top of the R-FCN method~\cite{RFCN}, which is a fully convolutional version of R-CNN framework among the state-of-the-art variants of R-CNN.  We implement our method using the latest MXNet. \textit{Our source code will be released.}
 
 \textbf{Setting and Implementation Details.} We conduct experiments with different settings: (i) Three different AOGs, $\mathcal{G}_{3,3,1}$, $\mathcal{G}_{5,5,1}$ and $\mathcal{G}_{7,7,2}$ ($\mathcal{G}_{7,7,1}$ is too slow to train, thus not reported). Note that we do not change the bounding box regression branch in the RoI prediction except for the RoI grid size which is changed to match the AOGs. (ii) Deformable vs non-deformable AOGs. We modified the latest R-FCN with deformable convolution~\cite{DCN} (i.e. RFCN-d) and we reused the code released on the Github~\footnote{\url{https://github.com/msracver/Deformable-ConvNets}}. For deformable AOGs, we allow Terminal-nodes deformable in computing their score vector $f_B(v)$'s similar to the deformable RoIPooling used in~\cite{DCN}. (iii) Folding vs Folding-Unfolding training procedure.  We follow the same hyper-parameter setting provided in the RFCN-d source code for fair comparison: the number of epochs is $7$, the learning rate starts with $0.0005$ and the scheduling step is at $4.83$, the warm-up step is used with a smaller learning rate $0.00005$ for $1000$ min-batches, and online hard-negative mining is adopted in training.

\textbf{Ablation Study.} The proposed integration of AOG and ConvNets is simple which substitutes the original RoIPooling operator with the AOGParsing operator. 
The RoIPooling is computed with a predefined and fixed flat configuration (e.g., $7\times 7$ grid). 
The AOGParsing is computed with a hierarchical and compositional AND-OR graph constructed on top of the same grid to explore much large number of latent configurations. Terminal-nodes use the same operators as the cells in the RoIPooling. AND-nodes and OR-nodes adopt very simple operators, SUM, MEAN or element-wise MAX. So, we expect that the proposed integration will not hurt the accuracy performance of the baseline R-CNN system, but is capable of output extractive rationale justification using the  parse trees inferred on-the-fly for each detected object. The RoIPooling operator is a special case of the AOGParsing operator. 
  
\begin{figure*} [!ht]
  \centering
  \includegraphics[width = 1\textwidth]{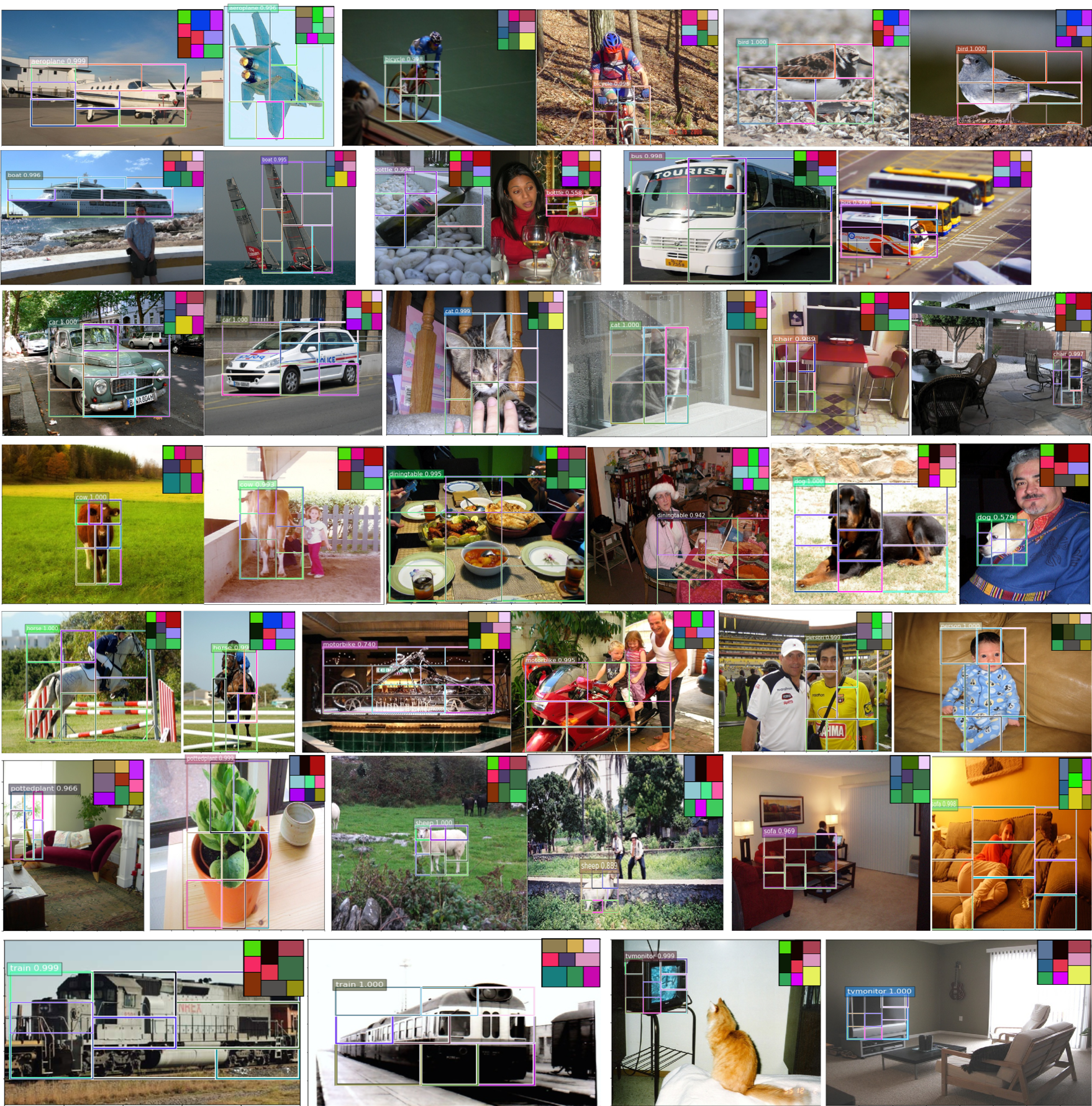}
  \caption{Examples of latent part configurations unfolded by AOGs using the learned model ``AOG772-1". We show one random example per category in VOC 2007 test dataset. We show one instance of the top two configurations for the 20 categories with the configuration superposed on the right-top of each image. (Best viewed in color and magnification) } \vspace{-3mm}
  \label{fig:results}
  \end{figure*}

We conduct ablation study on PASCAL VOC 2007. Table.~\ref{table:voc2007}  shows the breakdown performance and comparisons. The results show that all of the variants are comparable in terms of accuracy performance, which matches with our expectation. In terms of the extractive rationale justification, Figure~\ref{fig:results} shows some qualitative examples. 

   
\textbf{Results.} We also test the integration on the PASCAL VOC 2012 benchmark with results shown in  Table.~\ref{table:voc2012}. We report comparisons with the RFCN-d~\cite{RFCN,DCN} only since it is one of the state-of-the-art methods. 

\textbf{Runtime.} The runtime is mainly affected by the size of an AOG. Our current implementation of the AOG are not optimized with some operators are written in Python instead of C/C++. Per image, ``AOG772" roughly takes $0.78s$, ``AOG551" roughly takes $1.3s$ and ``AOG331" roughly takes $0.25s$. RFCN-d roughly takes $0.36s$. 

\textbf{Limitations and Discussions.} The proposed method has two main limitations to be addressed in future work. First, although it can show qualitative extractive rationales in detection in a weakly-supervised way, it is difficult to measure the model interpretability, especially in a quantitative way. For quantitative interpretability, we will investigate rigorous definitions which can be formalized as a interpretability-sensitive loss term in end-to-end training.  Second, current implementation of the proposed method did not improve the accuracy performance although it is not our focus in this paper. We will explore new operators for AND-nodes and OR-nodes in the AOG to improve performance. We hope detection performance will be further improved with the interpretability-sensitive loss term.

\section{Conclusion}\label{sec:conclusion}
This paper presented a method of integrating a generic top-down grammar model with bottom-up ConvNets in an end-to-end way for learning qualitatively interpretable models in object detection using the R-CNN framework. It builds on top the R-FCN method and substitutes the RoIPooling operator with an AOGParsing operator to unfold the space of latent part configurations. It proposed a folding-unfolding method in learning. In experiments, the proposed method is tested in the PASCAL VOC 2007 and 2012 benchmarks with performance comparable to state-of-the-art R-CNN based detection methods. The proposed method computes the optimal parse tree in the AOG as qualitatively extractive rationale in ``justifying" detection results. It sheds light on learning quantitatively interpretable models in object detection. 

\small{
\bibliographystyle{aaai}
\bibliography{reference}
}

\end{document}